\documentclass[10pt, a4paper]{article}

\usepackage[final]{lrec2026} 

\usepackage{graphicx}

\usepackage{booktabs}
\usepackage{multirow}
\usepackage{url}
\usepackage{subcaption}
\usepackage{float}

\usepackage{amsmath} 
\usepackage{makecell}
\usepackage[normalem]{ulem}

\title{Modeling the Human Lexicon under Temperature Variations: Linguistic Factors, Diversity and Typicality in LLM Word Associations}

\name{Maria A. Rodriguez$^{1, 3}$, Marie Candito$^{2}$, Richard Huyghe$^{1}$} 

\address{$^1$ University of Fribourg \\
    $^2$ LLF (Université Paris Cité / CNRS) \\
    $^3$ Lucerne University of Applied Sciences and Arts \\
         maria.anduezarodriguez@unifr.ch\\}

\abstract{
Large language models (LLMs) achieve impressive results in terms of fluency in text generation, yet the nature of their linguistic knowledge -- in particular the human-likeness of their internal lexicon -- remains uncertain. This study compares human and LLM-generated word associations to evaluate how accurately models capture human lexical patterns. Using English cue-response pairs from the SWOW dataset and newly generated associations from three LLMs (Mistral-7B, Llama-3.1-8B, and Qwen-2.5-32B) across multiple temperature settings, we examine (i) the influence of lexical factors such as word frequency and concreteness on cue-response pairs, and (ii) the variability and typicality of LLM responses compared to human responses. Results show that all models mirror human trends for frequency and concreteness but differ in response variability and typicality. Larger models such as Qwen tend to emulate a single ``prototypical'' human participant, generating highly typical but minimally variable responses, while smaller models such as Mistral and Llama produce more variable yet less typical responses. Temperature settings further influence this trade-off, with higher values increasing variability but decreasing typicality. These findings highlight both the similarities and differences between human and LLM lexicons, emphasizing the need to account for model size and temperature when probing LLM lexical representations.
\\ \newline 
}

\begin{document}

\maketitleabstract

\section{Introduction}

        The capacity of language models to generate texts of remarkable linguistic quality has spurred increasing interest in their underlying linguistic knowledge.
        Despite comparable abilities in written text production, human beings and language models differ fundamentally in the process of language acquisition: while human language learning is grounded in real-world experience and extralinguistic context, language models learn exclusively through large-scale exposure to text. This raises the question of whether these different learning processes result in comparable linguistic competence, understood as the implicit knowledge of linguistic systems and grammatical structures (see \citealp{waldis2024holmesbenchmarkassesslinguistic} for an overview).
        A particularly relevant domain for this comparison is the organization of the lexicon, paralleling research into the human mental lexicon. To explore lexical structures, recent investigations have examined the ability of large language models (LLMs) to generate word associations, comparing these with human word associations.
        These studies have employed diverse methodologies and yielded mixed results, revealing both shared characteristics and notable divergences between LLM- and human-generated word associations.
        In this paper, we contribute to this line of inquiry by examining the lexical factors that determine word associations in both groups and by proposing refined metrics for evaluating the human-likeness of associations generated by LLMs. 
        Our results provide new evidence of similarities between human and LLM word associations, while also showing that LLMs vary considerably in their ability to capture the diversity of human responses, with variability depending on models, settings, and cue words.

\section{Related work}
\label{sec:related}
        The word association task is a classic psychological experiment in which participants respond spontaneously with the first word(s) that come to mind when presented with a specific cue word (e.g., \textit{holiday} $\rightarrow$ \textit{vacation}, \textit{birth} $\rightarrow$ \textit{baby}, \textit{curly} $\rightarrow$ \textit{hair}). 
        Word associations have long been used by linguists and psychologists to investigate cognitive processes, psychological behavior patterns, language acquisition, multilingualism, as well as the organization of the mental lexicon (see \citealt{palermo1964word,cramer1968word, meara1983word,de2008word,Fitzpatrick2007WordAP}; a.o.).
       In particular, word association norms, assembled into networks linking cue and response words, have been shown to closely align with psycholinguistics results related to the structure of the mental lexicon -- as evidenced by analyses of the SWOW-EN dataset \cite{de2019small}, the largest existing collection of human-generated word associations in English.

        Given their capacity to reveal lexical organization, word associations provide a useful proxy for probing underlying lexical structures in LLMs and for comparing them with those of humans. The ability of LLMs to replicate human word associations has been explored in several recent studies.
        \citet{vintar2024human} investigated differences between human and model-generated word associations in Slovene and English. They fine-tuned encoder-decoder models to generate an open-ended list of associates given a cue. Based on examples derived from the English and Slovene SWOW datasets, the authors reported a low overlap between LLM and human responses. They further compared the types of responses provided by humans and LLMs, according to the classification into 4 categories (form, meaning, position, and erratic) and found a significantly higher proportion of erratic responses, i.e., responses apparently unconnected to cues, in LLMs. However, across relevant responses, they found similar distributions of meaning, form, and position (i.e., syntagmatic) relations between cues and responses.
        
        \citet{abramski2024,abramski2025llm} conducted a more direct comparison of human and LLM word associations by using similar prompts for LLMs as for humans (i.e., asking them to provide three response words for each cue). Based on the cue set from SWOW-EN, they created the LWOW datasets by eliciting English word associations from three LLMs (Mistral-7b, Llama3.1-8b, and Claude-3.5-Haiku-latest).
        The prompting was performed 100 times for each cue, and answers from these prompting repetitions were considered as parallels to answers from 100 human participants. The authors built 4 weighted networks from cue–response pairs -- one for SWOW-EN and one for each prompted LLM -- where edge weights reflected participant or prompt repetition. They validated the LLM networks using semantic priming data, showing that normalized spreading activation levels predicted lexical decision reaction times.
        Additionally, the authors conducted an exploratory analysis of the overlap between LLM and human responses, finding that human responses were considerably more diverse. This comparison was made between the sets of unique responses, irrespective of the associated cues.
        Unfortunately, the authors did not specify the temperature setting used in generating the LWOW datasets, although this hyperparameter may greatly impact the variability and human-likeliness of responses (see Section \ref{sec:typicality}).\footnote{The authors did not report the temperature parameter in their code and did not provide this information upon request. Our preliminary experiments using LWOW suggested that the default temperature of 1 was unlikely to have been applied. This led us to create new datasets to examine the effects of temperature variation.}

        \citet{xiao2025humanlikeness} also created datasets of LLM-generated word associations based on the SWOW-EN cues, but using larger models (GPT-4o and Llama-3.1-70b-instruct),\footnote{These datasets are not publicly available.} while focusing on evaluating whether LLMs capture core psycholinguistic dimensions of the mental lexicon.
        The authors found mixed results regarding the similarity between LLM- and human-derived networks. 
        They observed reduced alignment for models compared to humans, such as lower correlations between association frequency and reaction times in lexical decision and naming tasks, reduced lexical diversity, and differences in response distributions. Yet, correlations between random-walk relatedness scores and human judgments were nearly as high for human and model-derived networks.

        In the present study, we further investigate the ability of LLMs to generate human-like word associations. Rather than analyzing association networks, we focus directly on cue–response pairs and conduct two experiments using the SWOW-EN dataset. First, we examine some of the linguistic factors that are known to influence word associations, in particular word frequency and concreteness, and compare their influence across human and LLM word associations (Section \ref{sec:linguistic-factors}).
        We then assess the human-likeness of the responses provided by LLMs considering both their variability and typicality compared to human responses. To this end, we introduce and apply specific metrics to quantify the extent to which LLM responses are representative of human responses, moving beyond simple measures of response overlap between humans and models (Section \ref{sec:typicality}).

\section{Experiment 1: Linguistic factors of word associations}
\label{sec:linguistic-factors}

    In this section, we examine linguistic properties previously identified as influential in word associations, to compare their effects across associations produced by humans and LLMs. We focus on two key factors: word frequency and concreteness (see \citealt{rubin1986predicting,de1989representational,nelson1992word,nelson2000frequency,hill2014}; a.o.).\footnote{We also examined the effect of imageability (the extent to which a word readily evokes a mental image of its referent), a concept closely related to concreteness. As the results strongly paralleled those for concreteness, we do not discuss them in detail in this section.}
    Interestingly, \citet{abramski2024} reported that the ``concreteness effect" found in human responses -- namely that concrete cues elicit fewer distinct responses among participants -- was not observed in LLM responses.
    Our analysis concentrates on the relationship between cues and responses for both frequency and concreteness, comparing patterns between human and LLM-generated associations.

\subsection{Methodology}
    In this subsection, we present the datasets and metrics used in Experiment 1.
    
    \paragraph{Human word associations}
    We used word associations from SWOW-EN\footnote{https://smallworldofwords.org/en/project/home} \cite{de2019small}, which contains responses for 12,282 cues in English. These associations were collected by asking 100 participants to elicit the first 3 words that come to mind when presented with a cue word, resulting in over 150K unique cue-response pairs. Responses were ranked according to their order of occurrence (hereafter, R1, R2 and R3). However, in our experiments, we focus on R1 responses, as they reveal the most immediate and instinctive associations participants make with each cue.

    \paragraph{LLM word associations}
    We replicated the prompting procedure from LWOW \cite{abramski2025llm} to obtain datasets of LLM-generated word associations across various models and temperature settings. We used the same medium-size models as in LWOW: Mistral-7B-Instruct-v0.1 \cite{jiang2023mistral7b} and Llama3.1-8B-Instruct \cite{grattafiori2024llama3herdmodels}, although in their non-quantized version. However, for the larger model, we opted for the open-source Qwen2.5\footnote{More precisely, Qwen2.5-32B-Instruct-GPTQ-Int8, a quantized and instruction tuned version of Qwen2.5-32B.} \cite{Qwen2.5} instead of the closed-source Claude2.5-Haiku \cite{TheC3} used in LWOW. These models (hereafter Mistral, Llama, and Qwen) were prompted 100 times, each generating 3 responses per cue, replicating the SWOW protocol in which each cue is presented to 100 English speakers.\footnote{Note that prompting repetition is a proxy for considering a distribution of the most probable responses. It is a simplification meant to cope with subword tokenization and with closed models for which the internal distributions are not available.}
    Additionally, we controlled for temperature settings by generating responses at 4 distinct levels (0.3, 1, 1.5, and 2) for each model, resulting in 12 distinct LLM-derived datasets\footnote{https://github.com/mariro8/LLMWA}.

    Given the computational cost of querying a model 100 times for each cue, for 12 combinations of model and temperature, we decided to use only a subset of 500 cues uniformly sampled from the 12,282 cues in SWOW-EN -- which was reduced to 490 after filtering\footnote{We applied the post-processing procedure used in LWOW (e.g., lowercasing and spelling corrections) to enhance comparability. Additionally, we discarded cues  for which humans or models did not provide any response.}.
    We extracted from SWOW-EN the human responses to these 490 cues and verified that the subset was representative of the full dataset by comparing key metrics (see Section \ref{sec:typicality}).
    Ultimately, the procedure yielded 12 datasets of LLM-generated associations, each dataset containing 49K instances per response rank (R1, R2, R3). As in the human dataset, we focused on R1 responses to enable a direct comparison with human responses.

    \paragraph{Metrics}
    Word frequencies were extracted from a Wikipedia dump\footnote{https://github.com/IlyaSemenov/wikipedia-word-frequency} that comprises over 2.7M unique words -- considering only words that appear in at least 3 Wikipedia articles and do not contain digits.
    We assigned a frequency of 1 to cues and responses absent from this extraction.
    
    Information on word concreteness was retrieved from the English concreteness norms compiled by \citet{brysbaert2014concreteness}. These are based on ratings on a scale from 1 (abstract) to 5 (concrete) for approximately 40K words, collected from over 4K participants. When investigating concreteness, we only considered word associations with available concreteness scores for both the cue and the response in each (human and LLM) dataset.
    
    To examine the frequency and concreteness of responses in relation to their corresponding cues, we used relative measures of frequency and concreteness. These were calculated as the ratio of a response's frequency (or concreteness) to that of its cue.\footnote{For frequencies, we applied a base-2 logarithmic transformation to the ratio to compensate for the inherent skewness of the distributions.} These measures allowed us to assess whether a cue’s properties influence those of its responses, in both human and LLM-generated word associations.

\subsection{Results}

\paragraph{Frequency}

\begin{figure}[!tbp]
    \centering
    \includegraphics[width=0.48\textwidth]{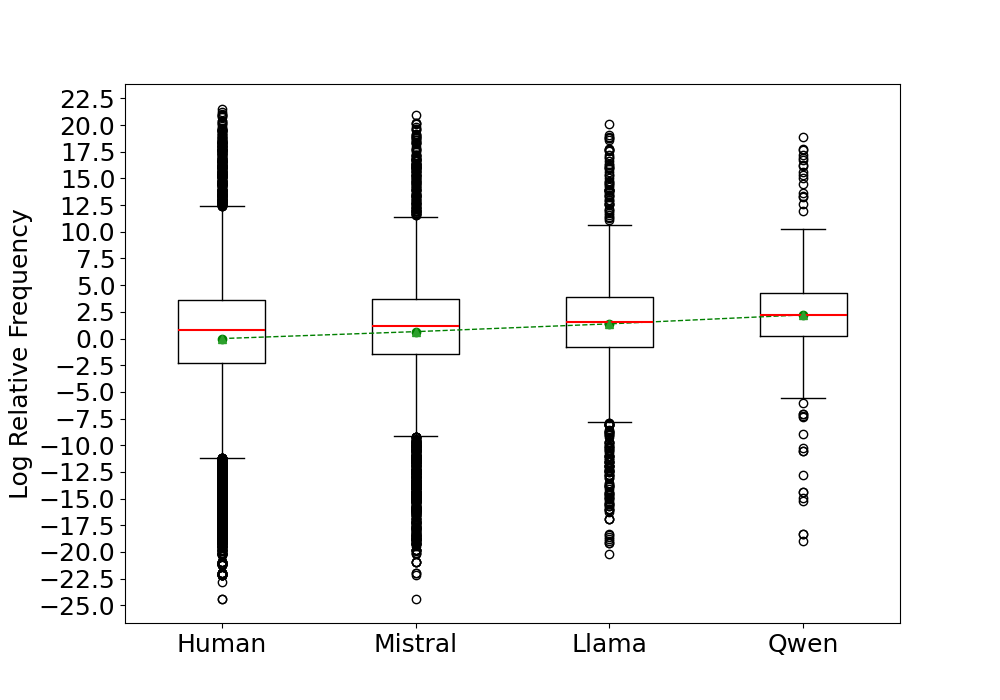}
    \caption{Distribution of log-transformed response/cue frequency ratios across respondents, considering responses provided by humans and LLMs at Rank 1, with LLM temperature set to 1.}
     \label{fig:rel_freq_R1.png}
\end{figure}

Figure \ref{fig:rel_freq_R1.png} presents the distribution of response-to-cue frequencies for cue-R1 unique pairs in both the human dataset and the three LLM datasets set at temperature 1.\footnote{We used a temperature of 1 as the default, as it likely corresponds to the training setting and therefore reflects the probability distributions learned by the models.} 
The differences in relative frequency distributions among the four respondent types are statistically significant according to a one-way ANOVA ($F$(3, 30,056) = 90.29, $p$~$<$~.001); however, the effect size is very small ($\eta^2$~=~.009), reflecting the modest extent of the differences observed between respondent types.

Response frequencies are on average of the same order of magnitude as cue frequencies across all respondents (humans and LLMs). More precisely, the mean response-to-cue frequencies are 0.00 for humans, and 0.65, 1.37, and 2.22 for Mistral, Llama, and Qwen, respectively. 
Some dispersion can be observed within each dataset, as the distribution of log-transformed frequency ratios is relatively broad. 
For human responses, for example, $Q1$~=~–2.3 and $Q3$~=~3.6, indicating that for a quarter of the cues, the response is approximately 5 times less frequent than the cue, while for another quarter, the response is 12 times more frequent than the cue. The dispersion is slightly lower for Mistral, Llama, and Qwen, in that order (Mistral: $Q1$~=~–1.4, $Q3$~=~3.7; Llama: $Q1$~=~–0.8, $Q3$~=~3.9; Qwen: $Q1$~=~0.2, $Q3$~=~4.3). 
Overall, cue and response frequencies show only a weak positive correlation ($r$~=~.16, $p$~$<$~.001).

\begin{figure}[!tbp]
    \centering
    \includegraphics[width=0.49\textwidth]{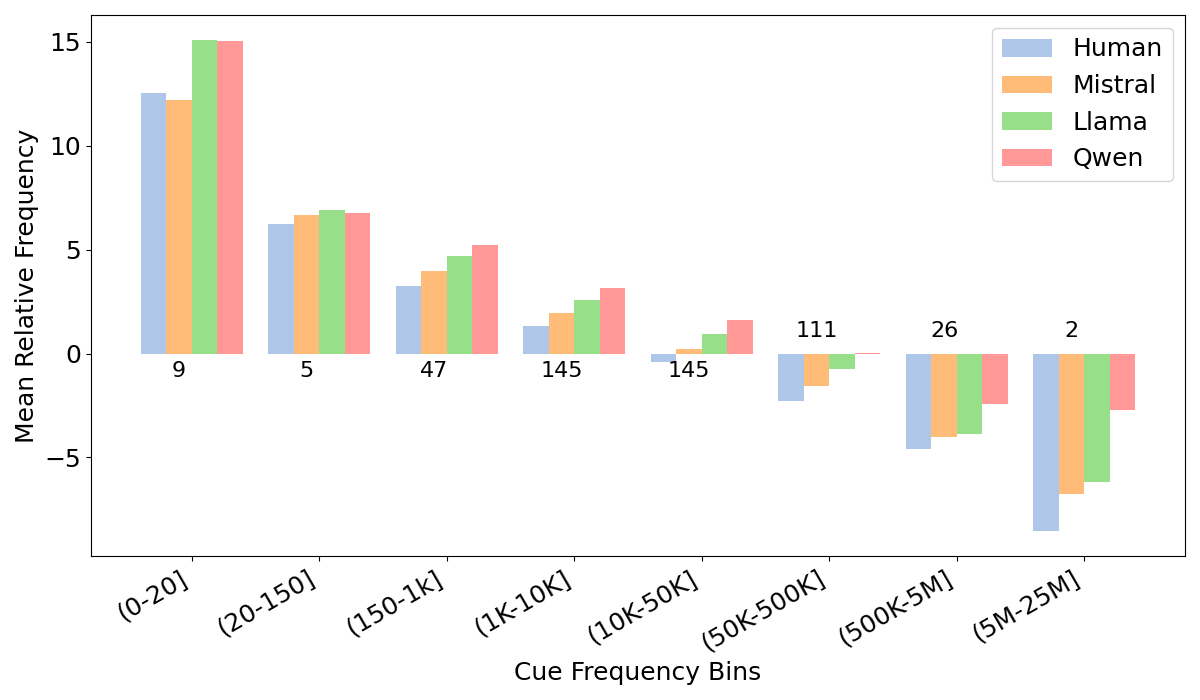}
    \caption{Average relative frequency of R1 responses across cue frequency bins (LLM temperature = 1). The number of cues included in each bin is indicated below or above the bars.}
    \label{fig:bins_rel_freq}
\end{figure}

To examine the relationship between cue and response frequencies in greater detail, we divided the set of cues into 10 bins of equal frequency range. The average relative frequencies across these bins are shown in Figure~\ref{fig:bins_rel_freq}. A consistent pattern emerges across all datasets: the lower the cue frequency, the higher the response-to-cue relative frequency. For approximately 75\% of the cues, ranging from rare to moderately frequent, responses tend to be more frequent than the cues. Conversely, the opposite trend is observed for frequent to highly frequent cues. This pattern is present in all 4 datasets, though it is generally more pronounced for humans, moderate for Mistral, and weaker for Llama and Qwen.

We investigated how response-to-cue relative frequencies in LLMs vary depending on temperature settings and found that temperature variation had a minimal impact.
Although ANOVA tests for each LLM indicated a statistically significant influence of temperature on frequency distributions ($p$ $<$ .001), the effect sizes were small ($\eta^2$ = [.007, .035]). In general, higher temperatures were associated with slightly lower relative frequencies (see Appendix \ref{appendix: rel_freq_temps}).

\begin{figure}[h]
    \centering
    \includegraphics[width=0.48\textwidth]{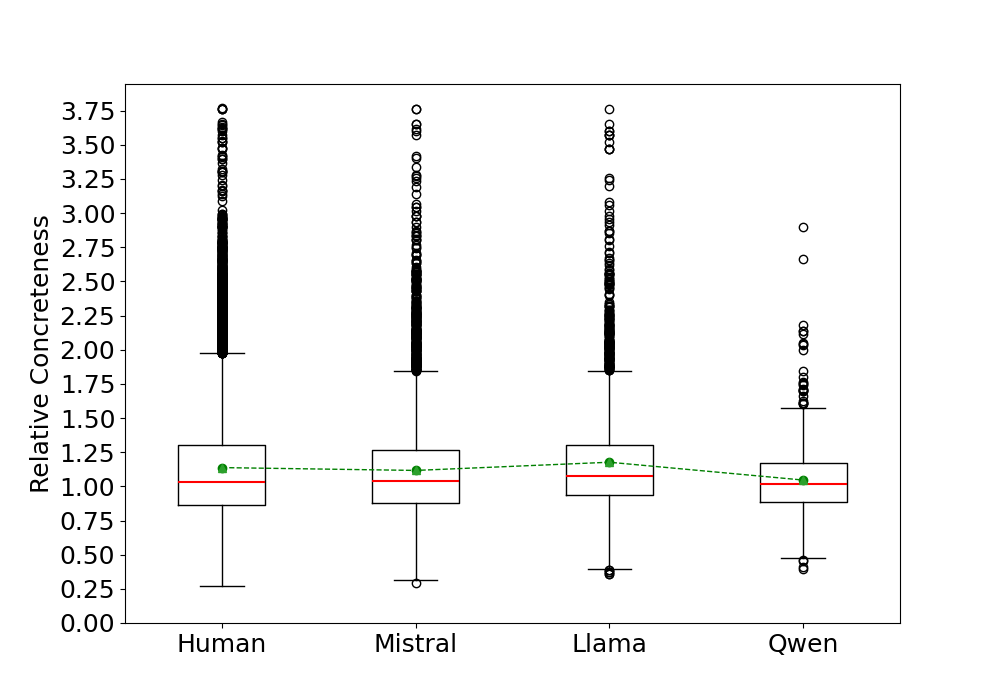}
    \caption{Distribution of response/cue concreteness ratios across respondents, considering responses provided by humans and LLMs at Rank 1, with LLM temperature set to 1.}
    \label{fig:rel_concr_R1.png}
\end{figure}

\paragraph{Concreteness}

Figure \ref{fig:rel_concr_R1.png} presents the distribution of the response-to-cue concreteness scores for cue-R1 unique pairs, in both the human dataset and the three LLM datasets at temperature 1. Humans and LLMs generally tend to elicit responses with the same level of concreteness as the cue, with only minor variation across respondent types: the mean relative concreteness values are 1.14 for humans, 1.12 for Mistral, 1.18 for Llama, and 1.05 for Qwen.
A one-way ANOVA indicates that differences in relative concreteness among the four respondents are statistically significant ($F(3, 24,885) = 21.89$, $p < .001$), but with a negligible effect size ($\eta^2$ = .003), consistent with the limited magnitude of these differences.

\begin{figure}[h]
    \centering
    \includegraphics[width=0.48\textwidth]{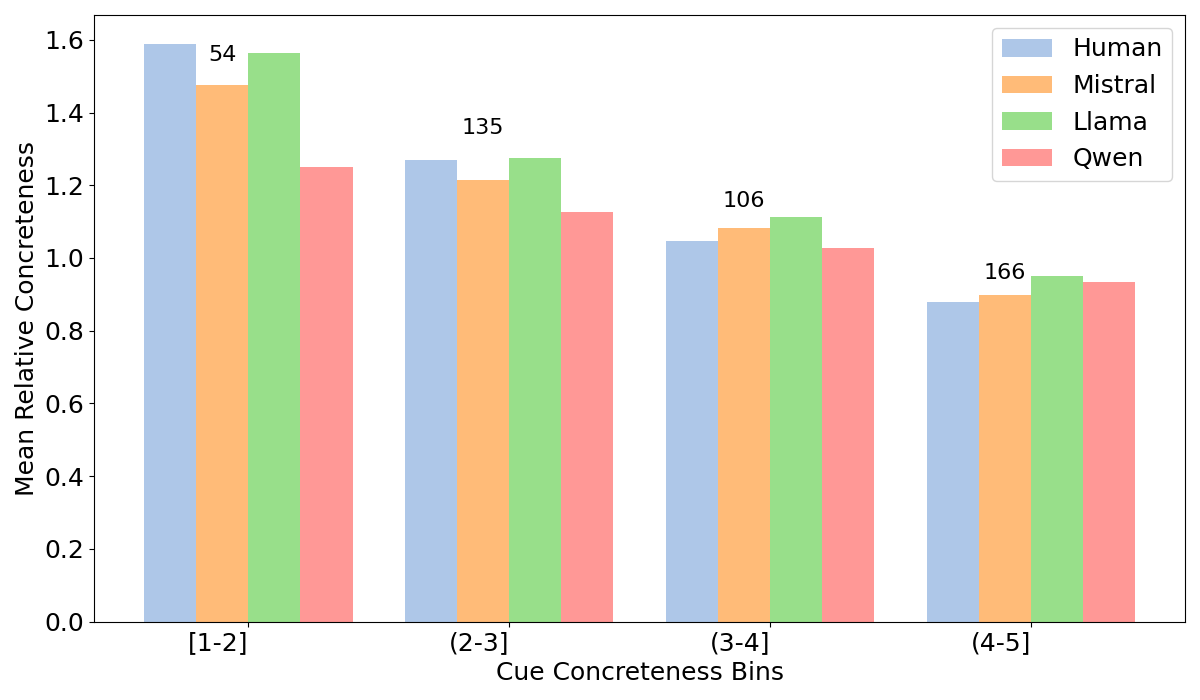}
    \caption{Average relative concreteness of R1 responses across cue concreteness bins (LLM temperature = 1). The number of cues included in each bin is indicated above the bars.}
    \label{fig:bins_rel_concr}
\end{figure}

A moderate correlation is found between cue and response concreteness ($r$ = .50, $p$ $<$ .001).
For all respondents, a small subset of responses are substantially more concrete than the cues. 
This can be explained by the tendency for abstract cues to elicit more concrete responses. 
When mean relative concreteness is broken down by cue concreteness bins, both humans and models show a systematic decline in relative concreteness as cue concreteness increases (see Figure \ref{fig:bins_rel_concr}). Responses to abstract cues are generally more concrete than the cue itself, but this effect diminishes progressively across bins, converging toward or below a ratio of 1 for highly concrete cues. Interestingly, while humans provide responses with higher relative concreteness for abstract cues, the models tend to generate responses with higher relative concreteness for concrete and very concrete cues.

We examined how the relative concreteness of LLM responses changes across temperature settings and found that temperature variation had almost no impact.
ANOVA tests revealed a statistically significant effect of temperature on the distribution of cue-to-response concreteness scores only for Mistral ($p < .001$), and even then the effect size was minimal ($\eta^2 = .003$) (see Appendix \ref{appendix: rel_concr_temps}).

\section{Experiment 2: Variability and typicality in word associations}
\label{sec:typicality}
    To assess the human-likeness of LLM word associations beyond structural factors such as frequency and concreteness, we examined the lexical content of individual responses. In this section, we compare human and LLM responses across two dimensions: variability (the number of distinct responses generated per cue) and typicality (the extent to which LLM word associations align with those frequently produced by humans for a cue).
    
    The number of distinct responses provided for a given cue word can vary considerably across datasets. For example, humans proposed 30 different responses for the cue word \textit{beach}, while Mistral, Llama and Qwen produced 17, 4 and 1 responses, respectively, at a temperature of 1.
    Responses may or may not be shared between humans and LLMs, and when they are, their frequency among respondents can differ substantially. A response that is common among humans -- and therefore highly typical of human responses -- may not appear as frequently in repeated responses generated by LLMs. Conversely, LLMs may frequently produce certain responses that are rarely provided by humans.
    For instance, in response to the cue word \textit{beach}, 47\% of humans responded with \textit{sand}, whereas the same response occurred in 49\% of Mistral's outputs, 31\% of Llama's, and 100\% of Qwen's. In this specific case, Mistral aligns more closely with human behavior than the two other LLMs.
    Since variability and typicality are conceptually independent, analyzing them separately allows for a nuanced comparison between human and LLM word associations.
    
\subsection {Methodology}
Experiment 2 used the same human and LLM datasets as Experiment 1. To study response variability in each dataset, we considered the number of distinct R1 responses per cue (hereafter $\#R1$) and examined its distribution across cues. 

To assess the typicality of LLMs responses, our starting point was the \textbf{associative strength} \cite{de2019small}, defined for a cue-response pair $(c,r)$ as the number of human participants having answered $r$ when given $c$, normalized for the total response tokens for $c$. 
Associative strength estimates the probability for a human 
to answer $r$ given $c$, $P(r | c)$, at a given rank.

Although associative strength provides an estimation of how salient a given response is in a group of participants for a given cue, its scale varies across cues as some of them elicit a few frequent responses while others generate many infrequent ones. 
To enable comparison of associative strengths across cues, we used a standardized measure defined as the deviation of a response’s strength from the average strength of all human responses for that cue (i.e., the $z$-score of human response strengths for that cue).
Let $D$ denote a set of word association instances, corresponding either to the human dataset (SWOW-EN) or to a set generated by a particular LLM at a specific temperature.
Let us note $(c, r, p, rk)\in D$ a word association instance within set $D$, where $r$ is the response to cue $c$ given by participant (or prompt repetition) $p$ at rank $rk$.
For a given set $D$, we use wildcards to denote subsets of $D$ -- e.g., $Llama_{0.3}(c,*,*,1)$ denotes the set of instances for $c$ provided at Rank 1 by Llama with temperature set to 0.3 (any response, any participant). Hence the associative strength of the $(c,r)$ word association, considering responses at Rank 1 is defined as
:
$$ S1(c,r) = \frac{|SWOW\text{-}EN(c,r,*,1)|}{|SWOW\text{-}EN(c,*,*,1)|}$$
Let $S1(c,*)$ be the distribution of S1 over all unique human responses for $c$. We define the {\bf standardized strength} at Rank 1 as:
\begin{equation}
 SS1(c,r) = \frac{S1(c,r) - \overline{S1(c,*)}}{SD(S1(c,*))}
 \label{eq:ss1}
 \end{equation}

Being derived from SWOW-EN, $SS1$ indicates the standardized strength of a response to a given cue in human word associations.
Nevertheless, 
one can also examine the $SS1$ of the $(c,r)$ pairs found in LLM datasets -- assigning a null strength $S1(c,r)$ to pairs that do not appear in SWOW-EN.
Accordingly, the $SS1$ of LLM responses can be used to evaluate how human-like the $(c,r)$ associations provided by LLMs are.

To obtain an overall measure of how typical of humans LLM responses are for a given cue, we averaged the standardized strengths of the LLM responses to that cue. For a given dataset $D$ of word association instances, we define the {\bf per-token average SS1} for a cue $c$ as: 
\begin{equation}
    tok\text{-}SS1_D(c) = \frac{\sum_{(c,r,p,1) \in D(c,*,*,1)}SS1(c,r)} {|D(c,*,*,1)|}
    \label{eq:d1_ss1}
\end{equation}
Equation \ref{eq:d1_ss1} averages over response tokens to account for the frequency of LLM responses across prompting repetitions. LLM responses are treated as a multiplicity of agents, and the formula measures how closely their responses align with typical human behavior.

In addition to the per-token average $SS1$, we also considered the $SS1$ averaged over the unique responses of a cue in a dataset $D$ -- hereafter the {\bf per-type average SS1}, ${typ\text{-}SS1_D}(c)$. This measure provides an evaluation of the human typicality of LLM responses independently of their number of repetitions.
Note that when considering the values of $typ\text{-}SS1_D(c)$ across the various LLMs, the reference point is 0, since averaging $SS1(c,r)$ over unique human responses in SWOW-EN is, by construction, equal to 0.
In contrast, there is no such inherent reference value for the $tok\text{-}SS1_D(c)$ in the human dataset. We only know that it must be greater than 0, as frequent human responses have positive standardized strengths. Accordingly, the value of $tok\text{-}SS1_{SWOW-EN}(c)$ serves as a calibration, indicating how skewed the distribution of human response tokens is for $c$ and providing a baseline for comparing the corresponding $tok\text{-}SS1_D(c)$ values in LLM datasets.

\subsection{Results}

Table \ref{tab:varia_and_typicality_res} presents the results for variability and typicality measures across human and LLM datasets at different temperature settings.
The table also provides information on how representative the 490-cue sample is of the full SWOW-EN cue set, showing that variability and typicality measures are highly similar between the sample and the complete human dataset (e.g., $avg\#R1$ of $38.4$ and $38.6$, respectively). This confirms the representativeness of the sample used in the experiments.

\begin{table}[ht]
        \centering
        \scalebox{0.675}{
        \begin{tabular}{|lc|rr|rr|}
        \cline{3-6}
        \multicolumn{2}{l|}{} & \multicolumn{2}{c|}{Variability} & \multicolumn{2}{c|}{Typicality} \\\hline
        \textbf{Resp.} & Temp. & $avg\#R1$ & $tot\#R1$ & $tok\text{-}SS1$ & $typ\text{-}SS1$\\
        \midrule
        \multirow{1}{*}{\textbf{Human}}
        & -      & \makecell[r]{38.4\\\{38.6\}} & \makecell[r]{9,175\\\{56,625\}} & \makecell[r]{1.67 (0.60)\\\{1.66 (0.57)\}}  & 0.00 (0.00)\\
        \midrule
        
        \multirow{4}{*}{\textbf{Mistral}}
        & 0.3       & 2.9     & 963    & 1.62 (1.89) & 1.36 (1.81) \\
        & 1.0       & 14.5    & 3,801  & 1.50 (1.38) & 0.14 (0.55) \\
        & 1.5       & 24.6    & 6,577  & 0.97 (1.15) & -0.18 (0.32) \\
        & 2.0       & 33.8    & 9,749  & 0.63 (0.87) & -0.34 (0.24)  \\
        
        \midrule
        
        \multirow{4}{*}{\textbf{Llama}}
        & 0.3       & 1.5    &   614  & 2.07 (2.11)       & 1.94 (2.15) \\
        & 1.0       & 6.8    & 2,202  & 1.81 (1.69)       & 1.06 (1.49) \\
        & 1.5       & 16.5   & 4,468  & 1.41 (1.37)       & 0.21 (0.81) \\
        & 2.0       & 27.5   & 6,996  & 1.00 (1.06)       & -0.20 (0.33) \\
        \midrule
        \multirow{4}{*}{\textbf{Qwen}}
        & 0.3    & 1.1 & 460   & 2.60 (2.11) & 2.54 (2.18) \\
        & 1.0    & 1.4 & 576   & 2.56 (2.03) & 2.45 (2.07) \\
        & 1.5    & 1.8 & 723   & 2.52 (1.95) & 2.33 (1.97) \\
        & 2.0    & 2.6 & 1,020 & 2.47 (1.83) & 2.14 (1.86) \\
        
        \bottomrule
        \end{tabular}
        }
        \caption{Measures of variability ($avg\#R1$: average number of unique R1 per cue; $tot\#R1$: total number of unique R1 responses) and typicality ($tok\text{-}SS1$ and $typ\text{-}SS1$, averaged over the 490 cues) for all respondents at different temperatures. Standard deviations are indicated in parentheses. Results for the entire set of SWOW-EN cues are presented in curly brackets.}
        \label{tab:varia_and_typicality_res}
    \end{table}

\begin{figure}[!tbp]
    \centering
    \includegraphics[width=0.49\textwidth]{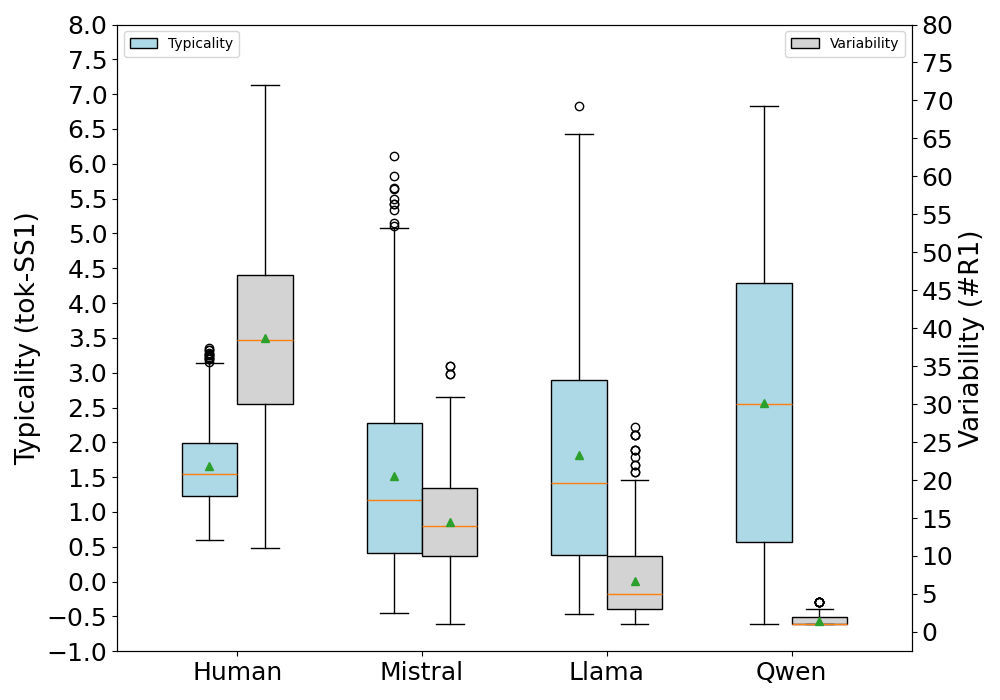}
    \caption{Distribution of variability ($\#R1$) and typicality ($tok\text{-}SS1$) in human and LLM datasets at temperature 1. Grey boxplots (right y-axis) represent variability, while blue boxplots (left y-axis) represent typicality.}
    \label{fig:R1_SS1_and_varia}
\end{figure}

\paragraph{Results at temperature 1}
 We first examined the results obtained with models prompted at temperature 1. 
Human responses appeared more diverse than those of all three LLMs, with an average $\#R1$ of 38.4 for humans, compared to 14.5, 6.8, and 1.4 for Mistral, Llama, and Qwen, respectively. Variability itself differs across models, with average $\#R1$ values progressively decreasing from Mistral to Llama to Qwen. As shown in Figure \ref{fig:R1_SS1_and_varia}, this decrease in average variability is accompanied by an increasingly concentrated distribution over cues. In particular, Qwen at temperature 1 exhibits very limited variability of R1 responses -- less than 5 distinct responses per cue for all cues -- whereas for Mistral, three quarters of cues elicit more than 10 distinct responses. 

As far as typicality is concerned, the average reference value of human $tok\text{-}SS1$ across the 490 cues is 1.67 (see Table \ref{tab:varia_and_typicality_res}). The models at temperature 1 obtain a comparable or much higher $tok\text{-}SS1$ value (Mistral: 1.50; Llama: 1.81; Qwen: 2.56). Thus, for Qwen and to a lesser extent Llama, the R1 responses tend to correspond to those frequently given by human participants. This result may be explained by model size, particularly for Qwen (32B, quantized), which has substantially more parameters than Llama (8B) and Mistral (7B).
That said, a closer examination of the complete distributions (Figure \ref{fig:R1_SS1_and_varia}) shows that human responses exhibit the least dispersion of $tok\text{-}SS1$ across cues. Furthermore, a higher average $tok\text{-}SS1$ in LLMs is accompanied by greater dispersion across cues: for Qwen especially, some cues elicit responses that are highly typical of humans ($Q3$~=~4.5), while negative $tok\text{-}SS1$ values are observed for all three models, indicating that for certain cues, LLM responses poorly align with human data.

Overall, variability and typicality tend to be negatively correlated in LLMs ($r$~=~–.42, $p$~$<$~.001). As illustrated in Figure \ref{fig:R1_SS1_and_varia}, models generating less variable responses generally produce more human-typical associations. This means that the responses with high model-assigned probabilities (hence repeated across the 100 repetitions) tend to correspond to more human-typical responses. 
Qwen, in particular, seems to behave like a single agent generating the associations most commonly produced by multiple human participants.

\begin{figure}[!tbp]
    \centering
    \includegraphics[width=0.489\textwidth]{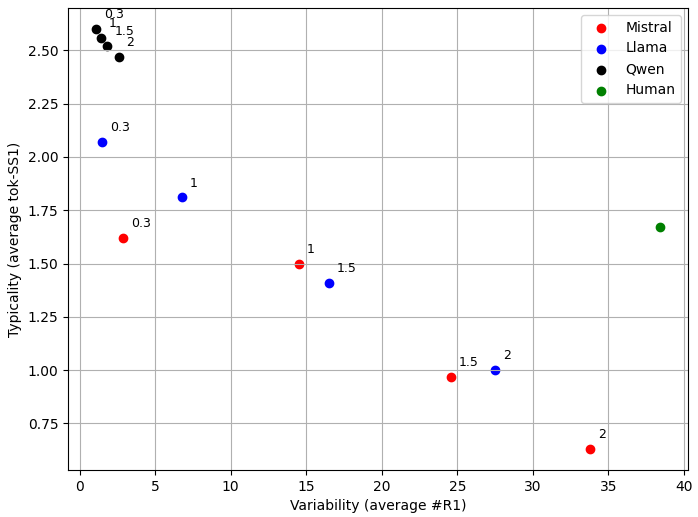}
    \caption{Variability ($\#R1$) vs. typicality ($tok\text{-}SS1$), both averaged over the 490 cues, for all respondents and temperatures. Temperatures are indicated as dot labels.}
    \label{fig:typicality_vs_variability}
\end{figure}

\paragraph{Impact of temperature variation}
In addition to the results obtained at the default temperature of 1, we analyzed how altering the sampling temperature affect both variability and typicality. The impact of these temperature changes is illustrated in Figure~\ref{fig:typicality_vs_variability}.
The primary finding regarding variability -- that it increases with higher temperature -- is unsurprising. However, two additional points are noteworthy. First, across all models and temperatures tested, variability remains below that observed in a sample of 100 human participants. At a temperature of 2, the average $\#R1$ is 33.8, 27.5, and 2.6 for Mistral, Llama, and Qwen, respectively, compared to 38.4 for humans -- for Mistral and Llama, further increases in temperature would likely push variability above human levels. Second, the effect of temperature on variability differs across models. For the largest model, Qwen, response variability remains very low even at temperature 2, indicating that the model is extremely “confident” in its top-1 responses (with a logit score far exceeding the others). This suggests that Qwen better represents a single prototypical agent than a heterogeneous population.

Unlike variability, typicality generally decreases with higher temperature:
both per-token and per-type typicality scores decline as temperature increases. It seems that enhancing the probability of selecting responses with lower logit scores comes at the cost of providing typically human answers.

For $typ\text{-}SS1$, the majority of cue-averaged values are positive across the three LLMs (see Table \ref{tab:varia_and_typicality_res}), with the exceptions of Llama at temperature 2 and Mistral at temperatures 1.5 and 2. It appears that, in most cases, the models tend to generate responses that are not only provided by humans but are also given by a greater-than-average number of human participants. This effect is particularly pronounced for Qwen: across all tested temperatures, the mean $typ\text{-}SS1$ ranges from 2.14 to 2.54. In other words, Qwen's responses to a cue correspond to those given by a number of human participants that is 2.14 to 2.54 standard deviations above the mean for that cue. Nevertheless, substantial variation of $typ\text{-}SS1$ is observed among cues (with $SD$ ranging from 1.86 to 2.18 across temperatures), meaning that the trend is stronger for certain cues and weaker for others. In contrast, for Mistral and Llama, average $typ\text{-}SS1$ can turn negative at certain temperatures, indicating a tendency to generate responses that are rarely or never provided by humans.

Turning to $tok\text{-}SS1$, the human data provide the reference values ($M$~=~1.66, $SD$~=~0.57).
All $tok\text{-}SS1$ values are higher than the corresponding $typ\text{-}SS1$ values. While this trend ($tok\text{-}SS1 > typ\text{-}SS1$) holds by construction for the human dataset, it is not guaranteed a priori for the models. The fact that it emerges consistently in the LLM datasets shows that the LLM responses with high human strength are also those that the models tend to produce repeatedly across the 100 trials.

The negative correlation between variability and typicality observed across models at temperature 1 persists when temperature is varied. Since variability increases with higher temperature, lower temperatures are associated with both reduced variability and increased typicality, as can be seen in Figure~\ref{fig:typicality_vs_variability}. In other words, responses with the highest logit scores -- which become increasingly likely to be provided by LLMs as temperature decreases -- tend to be those that are most typical of humans. However, the effect of temperature varies across models. Between temperatures 0.3 and 2, $tok\text{-}SS1$ decreases by a factor of 2.6 for Mistral, by roughly a factor of 2 for Llama, and only minimally for Qwen.
Furthermore, when comparing models at similar levels of variability, a larger model like Qwen reaches much higher typicality than smaller models (see, e.g., Qwen at temperature 2 vs. Mistral at temperature 0.3 in Table \ref{tab:varia_and_typicality_res} and in the left part of Figure \ref{fig:typicality_vs_variability}). This suggests that Qwen's top-scoring responses are substantially more human-typical than those produced by the medium-sized models tested.

The main conclusion that can be drawn from these observations is that temperature affects both variability and typicality in LLM word associations. While the effect on variability directly follows from the definition of temperature, its impact on typicality is less predictable. Our findings indicate that the responses with the highest probability in LLMs also tend to be highly typical of humans. These results highlight the overall importance of treating temperature as a critical factor in research examining lexical organization in LLMs.

\section{Discussion and conclusion}



In this study, we explored the ability of LLMs to generate word associations that are comparable to those produced by humans. Using datasets containing human and LLM responses to the same English cue words, we analyzed the structural properties of LLM associations and found that lexical factors influencing human associations exert similar effects on LLM-generated ones. In particular, the frequency and concreteness of the responses relative to the cues were closely aligned between the two sources. This suggests that LLMs have effectively internalized certain lexical mechanisms underlying word associations, as well as some of the lexical relationships that structure the human mental lexicon. Such properties may account for the similarity between lexical networks derived from LLM associations and those based on human data, as observed by \citet{abramski2025llm} and, to a lesser degree, by \citet{xiao2025humanlikeness}.

To further examine the lexical content of associations and the characteristics of individual responses, we assessed how effectively LLM generations reflect the recurrence and diversity observed in human data.
Based on the associative strength of cue-response pairs -- the proportion of human participants who provided a particular response to a given cue --, we developed metrics to quantify how typical of humans a set of LLM responses is for a cue.
Our analysis revealed that LLMs may vary in their capacity to reflect human associations, yet none fully replicates the patterns of typicality and variability observed in human responses.

On the one hand, the ability of LLMs to reproduce human associations depends on model settings and hyperparameters. Temperature, in particular, plays a crucial role, which had not been previously explored in studies on LLM-generated word associations. Higher temperatures tend to increase variability while reducing typicality. Accordingly, depending on temperature, LLMs can either approximate the typicality of human responses at the expense of variability, or vice versa. Nevertheless, none of the tested LLMs was able to match the balance between lexical variation and recurrence found in human word associations.

On the other hand, notable differences emerged among LLMs, independent of temperature effects. While the largest model we tested (Qwen2.5-32B) shows a general preference for representing a single prototypical human who consistently produces the most common responses, medium-sized models (Mistral-7b and Llama3.1-8b) more closely resemble a group of individuals with potentially idiosyncratic responses -- although these do not fully align with human associations. These differences appear to be influenced by model size, with larger LLMs tending to exhibit greater uniformity, prototypicality and single-agent behavior than smaller models. Moreover, important variation in response typicality was observed across cues, even in models that generally produce the most typical responses, indicating that human associations for certain cues still resist accurate modeling.

These findings provide further evidence that the lexical knowledge of LLMs differs in nature from that of humans, and complement the contrasting trends observed by \citet{xiao2025humanlikeness}. Because of differences in language acquisition and in the volume of text they are exposed to, LLMs and humans build their lexicons in fundamentally different ways. A key difference is that an LLM may be capable of representing multiple mental lexicons -- which is not natural for a human subject -- whereas the notion of a ``prototypical'' human speaker is itself an abstraction that is not embodied by any single individual. Nevertheless, capturing human-specific lexical idiosyncrasies remains a challenge for LLMs and requires additional adaptations. Exploring how accurately an LLM can reflect the diversity of human mental lexicons offers a promising direction for future research on the linguistic capabilities of artificial intelligence.



\section{Limitations}
This study has several limitations, three of which are worth mentioning.
First, our analysis was restricted to open-source models (Mistral-7B, Llama-3.1-8B, and Qwen-2.5-32B). While this choice may improve transparency and reproducibility, it excludes closed-source models such as Claude-3.5-Haiku (used by \citealt{abramski2024}), GPT-4o (used by \citealt{xiao2025humanlikeness}), or even larger models, which may show different lexical behaviors. 

Second, due to computational and environmental constraints, we used only 490 cues from the 12,000+ cues in the SWOW-EN dataset. 
Although we verified that this subset was representative in terms of variability and typicality, using the full dataset might reveal finer-grained effects than those highlighted in the present study.

Third, our evaluation of human-likeness relied on quantitative metrics (frequency, concreteness, typicality, variability). While these capture important lexical dimensions, they may not fully reflect the qualitative nuances of human word associations. Future work should combine quantitative assessment with a more qualitative evaluation to finely analyze the human-likeness of LLM associations. Moreover, the dispersion observed in typicality and variability across cues is quite significant and calls for a detailed investigation.
A complete comparison of human and LLM-generated word associations requires further examination of both the cues that trigger typical human responses and the human responses that LLMs do or do not reproduce.

\section{Bibliographical References}\label{sec:reference}

\bibliographystyle{lrec2026-natbib}
\bibliography{custom}

@misc{qwen2.5,
    title = {Qwen2.5: A Party of Foundation Models},
    url = {https://qwenlm.github.io/blog/qwen2.5/},
    author = {Qwen Team},
    month = {September},
    year = {2024}
}

@article{nelson1992word,
  title={Word concreteness and word structure as independent determinants of recall},
  author={Nelson, Douglas L and Schreiber, Thomas A},
  journal={Journal of memory and language},
  volume={31},
  number={2},
  pages={237--260},
  year={1992},
  publisher={Elsevier}
}

@article{de1989representational,
  title={Representational aspects of word imageability and word frequency as assessed through word association.},
  author={de Groot, Annette M},
  journal={Journal of Experimental Psychology: Learning, Memory, and Cognition},
  volume={15},
  number={5},
  pages={824},
  year={1989},
  publisher={American Psychological Association}
}

@article{rubin1986predicting,
  title={Predicting which words get recalled: Measures of free recall, availability, goodness, emotionality, and pronunciability for 925 nouns},
  author={Rubin, David C and Friendly, Michael},
  journal={Memory \& cognition},
  volume={14},
  number={1},
  pages={79--94},
  year={1986},
  publisher={Springer}
}

@inproceedings{vintar2024human,
  title={How Human-Like are Word Associations in Generative Models? An Experiment in Slovene},
  author={Vintar, {\v{S}}pela and Brglez, Mojca and {\v{Z}}agar, Ale{\v{s}}},
  booktitle={Proceedings of the Workshop on Cognitive Aspects of the Lexicon@ LREC-COLING 2024},
  pages={42--48},
  year={2024}
}

@article{meara1983word,
  title={Word associations in a foreign language},
  author={Meara, Paul},
  journal={Nottingham Linguistics Circular},
  volume={11},
  number={2},
  pages={29--38},
  year={1983}
}

@article{nelson2000frequency,
title={What is this think called frequency?},
author={Nelson, Douglas L and McEvoy, Cathy L},
year=2000,
pages={509--522},
volume=28,
number={4},
  journal={Memory \& Cognition},
  publisher={Springer}
}

@article{Fitzpatrick2007WordAP,
  title={Word association patterns: unpacking the assumptions},
  author={Tess Fitzpatrick},
  journal={International Journal of Applied Linguistics},
  year={2007},
  volume={17},
  pages={319-331},
}

@article{de2008word,
  title={Word associations: Network and semantic properties},
  author={De Deyne, Simon and Storms, Gert},
  journal={Behavior research methods},
  volume={40},
  number={1},
  pages={213--231},
  year={2008},
  publisher={Springer}
}

@article{brysbaert2014concreteness,
  title={Concreteness ratings for 40 thousand generally known English word lemmas},
  author={Brysbaert, Marc and Warriner, Amy Beth and Kuperman, Victor},
  journal={Behavior research methods},
  volume={46},
  number={3},
  pages={904--911},
  year={2014},
  publisher={Springer}
}

@article{de2019small,
  title={The “Small World of Words” English word association norms for over 12,000 cue words},
  author={De Deyne, Simon and Navarro, Danielle J and Perfors, Amy and Brysbaert, Marc and Storms, Gert},
  journal={Behavior research methods},
  volume={51},
  pages={987--1006},
  year={2019},
  publisher={Springer}
}

@misc{jiang2023mistral7b,
      title={Mistral 7B}, 
      author={Albert Q. Jiang and Alexandre Sablayrolles and Arthur Mensch and Chris Bamford and Devendra Singh Chaplot and Diego de las Casas and Florian Bressand and Gianna Lengyel and Guillaume Lample and Lucile Saulnier and Lélio Renard Lavaud and Marie-Anne Lachaux and Pierre Stock and Teven Le Scao and Thibaut Lavril and Thomas Wang and Timothée Lacroix and William El Sayed},
      year={2023},
      eprint={2310.06825},
      archivePrefix={arXiv},
      primaryClass={cs.CL},
      url={https://arxiv.org/abs/2310.06825}, 
}

@misc{grattafiori2024llama3herdmodels,
      title={The Llama 3 Herd of Models}, 
      author={Aaron Grattafiori and Abhimanyu Dubey and Abhinav Jauhri and Abhinav Pandey and Abhishek Kadian and others},
      year={2024},
      eprint={2407.21783},
      archivePrefix={arXiv},
      primaryClass={cs.AI},
      url={https://arxiv.org/abs/2407.21783}, 
}

@misc{TheC3,
  title={The Claude 3 Model Family: Opus, Sonnet, Haiku},
  author={Anthropic},
  year={2024},
  url={https://api.semanticscholar.org/CorpusID:268232499}
}

@article{abramski2025llm,
  title={The “LLM World of Words” English free association norms generated by large language models},
  author={Abramski, Katherine and Improta, Riccardo and Rossetti, Giulio and Stella, Massimo},
  journal={Scientific data},
  volume={12},
  number={1},
  pages={803},
  year={2025},
  publisher={Nature Publishing Group UK London}
}

@inproceedings{abramski2024,
  title={"LLM-Generated Word Association Norms"},
author={Abramski, Katherine and Lavorati, Clara and Rossetti, Giulio and Stella, Massimo},
booktitle={Proceedings of the Third International Conference on Hybrid Human-Artificial Intelligence (HHAI)},
pages={3--13},
year={2024},
address={Malm{ö}, Sweden},
}

@inproceedings{
xiao2025humanlikeness,
title={Human-likeness of {LLM}s in the Mental Lexicon},
author={Bei Xiao and Xufeng Duan and David A. Haslett and Zhenguang Cai},
booktitle={to appear in the proceedings of the SIGNLL Conference on Computational Natural Language Learning (CoNLL 2025)},
year={2025},
url={https://openreview.net/forum?id=beu7HZAYtG}
}

@article{hill2014,
author = {Hill, Felix and Korhonen, Anna and Bentz, Christian},
title = {A Quantitative Empirical Analysis of the Abstract/Concrete Distinction},
journal = {Cognitive Science},
volume = {38},
number = {1},
pages = {162-177},
doi = {https://doi.org/10.1111/cogs.12076},
url = {https://onlinelibrary.wiley.com/doi/abs/10.1111/cogs.12076},
eprint = {https://onlinelibrary.wiley.com/doi/pdf/10.1111/cogs.12076},
year = {2014}
}

@misc{waldis2024holmesbenchmarkassesslinguistic,
      title={Holmes: A Benchmark to Assess the Linguistic Competence of Language Models}, 
      author={Andreas Waldis and Yotam Perlitz and Leshem Choshen and Yufang Hou and Iryna Gurevych},
      year={2024},
      eprint={2404.18923},
      archivePrefix={arXiv},
      primaryClass={cs.CL},
      url={https://arxiv.org/abs/2404.18923}, 
}

@book{cramer1968word,
  title={Word Association},
  author={Cramer, P.},
  isbn={9780121964504},
  lccn={68014652},
  series={Word Association},
  url={https://books.google.ch/books?id=WV1-AAAAMAAJ},
  year={1968},
  publisher={Academic Press}
}

@book{palermo1964word,
  title={Word association norms: Grade school through college.},
  author={Palermo, David S and Jenkins, James J},
  year={1964},
  publisher={U. Minnesota Press}
}

\clearpage
\appendix

\section{Appendix A: Temperature effects}
\subsection{Relative frequency}
\label{appendix: rel_freq_temps}
\vspace{-25pt}
\begin{figure}[H]
    \centering
    \includegraphics[width=0.45\textwidth]{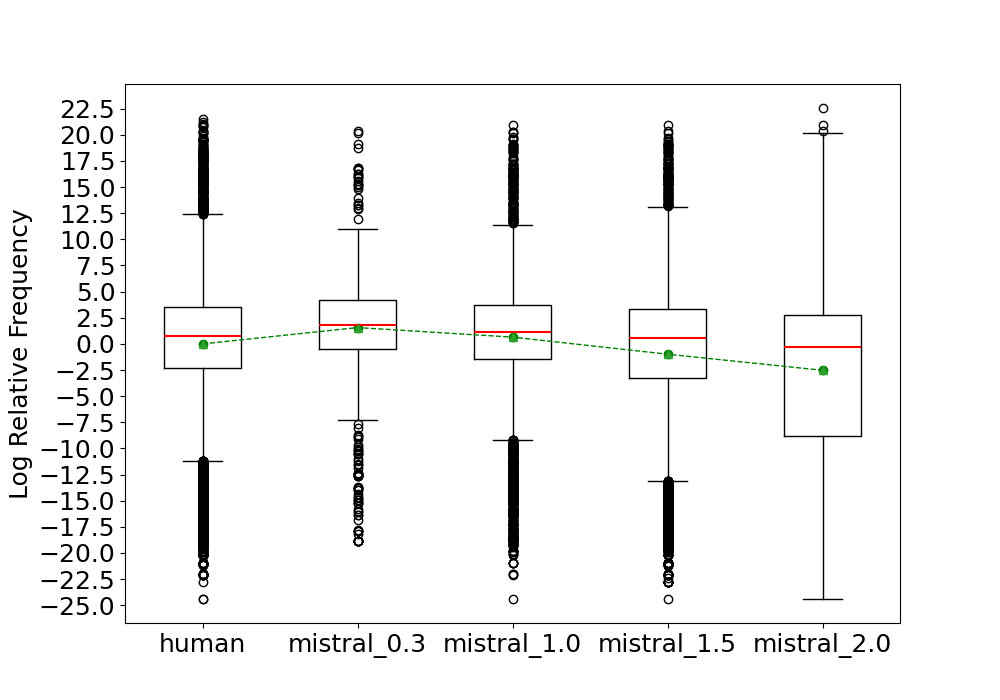}
    \caption{Mistral: log relative frequency across temperatures w.r.t. humans.}
    \label{fig:mistral_rel_freq_temps}
\end{figure}
\vspace{-40pt}

\begin{figure}[H]
    \centering
    \includegraphics[width=0.45\textwidth]{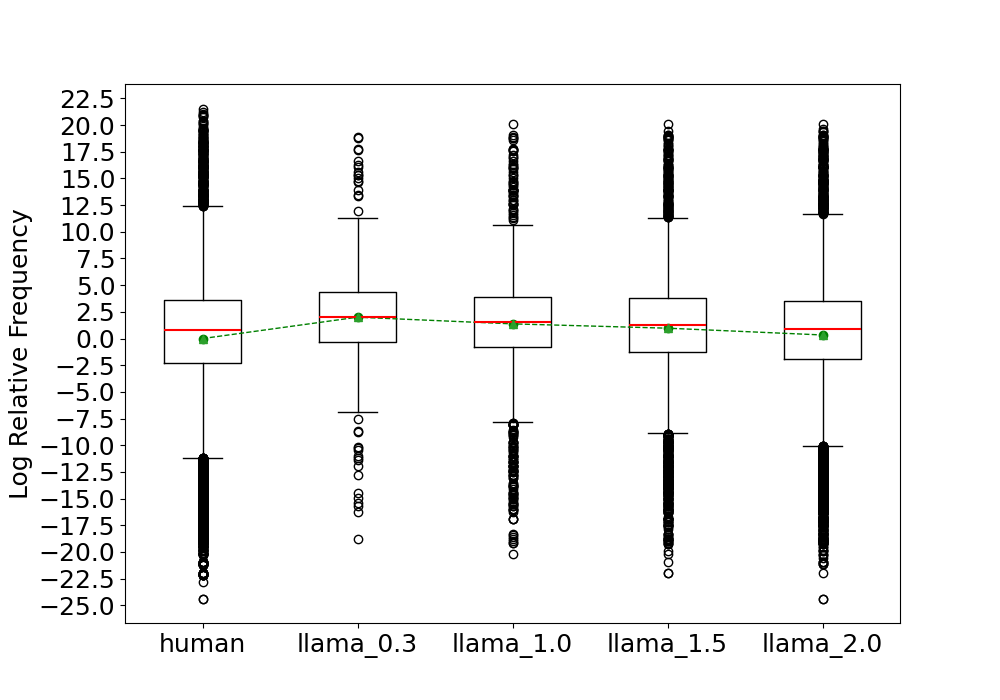}
    \caption{Llama: log relative frequency across temperatures w.r.t. humans.}
    \label{fig:llama_rel_freq_temps}
\end{figure}
\vspace{-40pt}

\begin{figure}[H]
    \centering
    \includegraphics[width=0.45\textwidth]{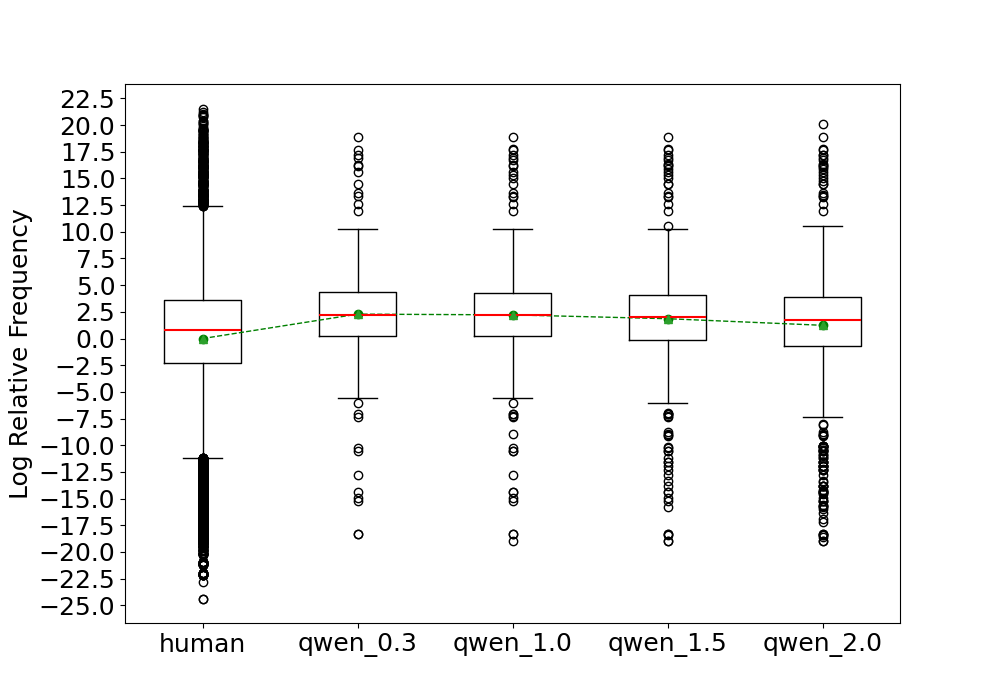}
    \caption{Qwen: log relative frequency across temperatures w.r.t. humans.}
    \label{fig:qwen_rel_freq_temps}
\end{figure}

\begin{table}[H]
    \centering
    \begin{tabular}{lccc}
        \toprule
        Model & $F$ & $p$ & $\eta^2$ \\
        \midrule
        Mistral & 449.82 & .000 & .035 \\
        Llama   & 64.54 & .000 & .007 \\
        Qwen    & 10.21 & .000 & .009 \\
        \bottomrule
    \end{tabular}
    \caption{ANOVA results for the effect of temperature on relative frequencies.}
    \label{tab:appendix_freq_anova}
\end{table}

\subsection{Relative concreteness}
\label{appendix: rel_concr_temps}
\vspace{-20pt}

\begin{figure}[H]
    \centering
    \includegraphics[width=0.45\textwidth]{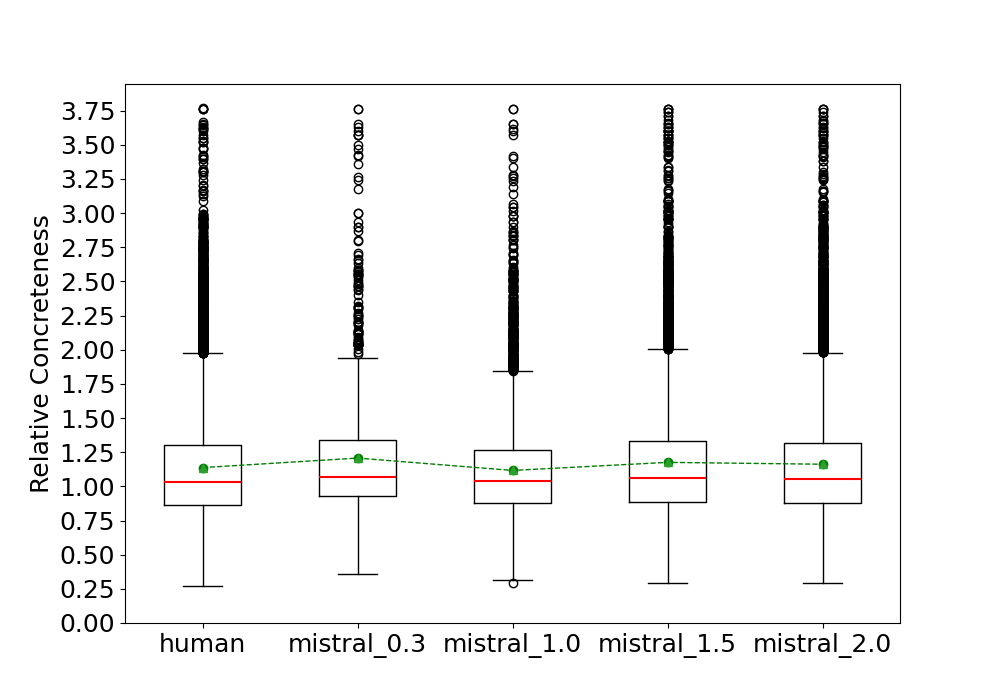}
    \caption{Mistral: relative concreteness across temperatures w.r.t. humans.}
    \label{fig:mistral_rel_concr_temps}
\end{figure}
\vspace{-25pt}

\begin{figure}[H]
    \centering
    \includegraphics[width=0.45\textwidth]{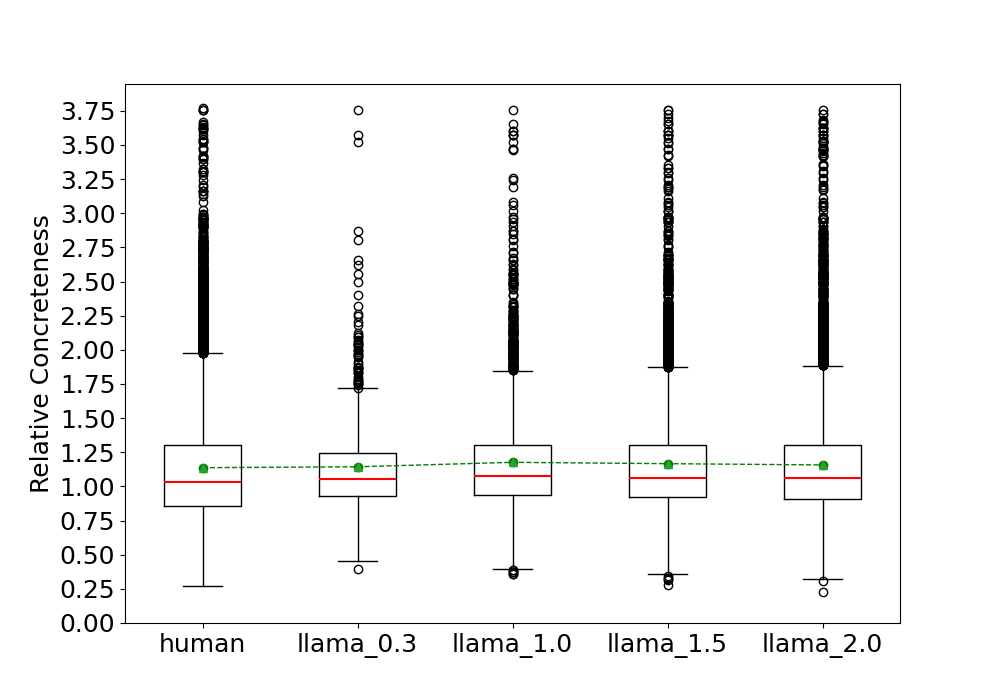}
    \caption{Llama: relative concreteness across temperatures w.r.t. humans.}
    \label{fig:llama_rel_concr_temps}
\end{figure}
\vspace{-25pt}

\begin{figure}[H]
    \centering
    \includegraphics[width=0.45\textwidth]{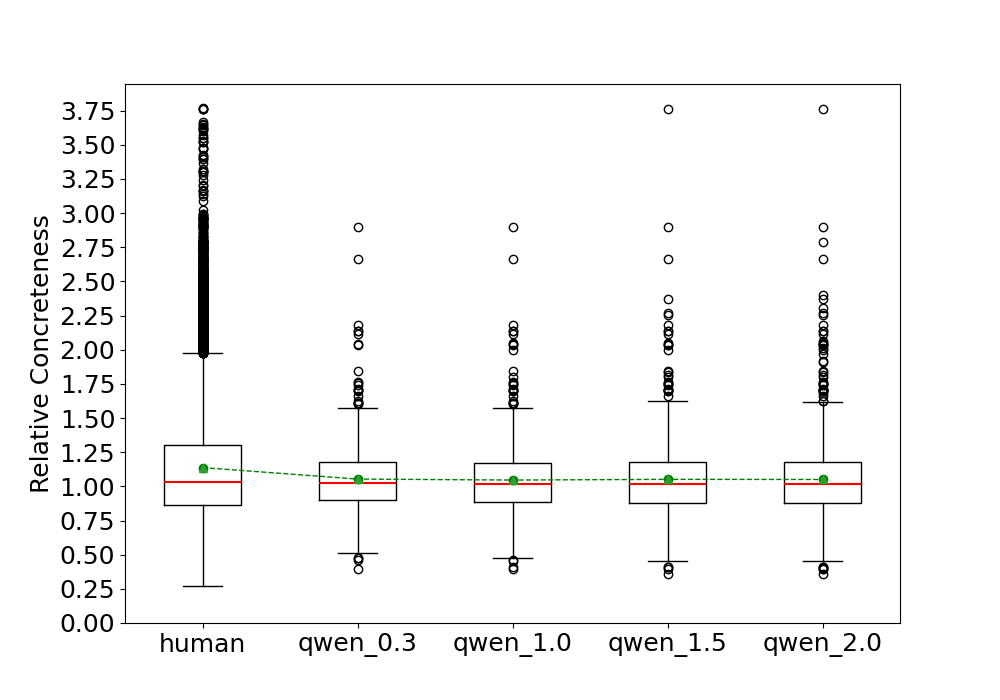}
    \caption{Qwen: relative concreteness across temperatures w.r.t. humans.}
    \label{fig:qwen_rel_concr_temps}
\end{figure}

\begin{table}[H]
    \centering
    \begin{tabular}{lccc}
        \toprule
        Model & $F$ & $p$ & $\eta^2$ \\
        \midrule
        Mistral & 26.17 & .000 & .003 \\
        Llama   & 2.26 & .079 & - \\
        Qwen    & .06 & .980 & - \\
        \bottomrule
    \end{tabular}
    \caption{ANOVA results for the effect of temperature on relative concreteness.}
    \label{tab:appendix_concr_anova}
\end{table}

\end{document}